\documentclass[article,12pt]{elsarticle}

\usepackage{amssymb}

\usepackage{siunitx}
\usepackage{graphicx}
\usepackage{hyperref}
\usepackage{alltt}
\usepackage{longtable}

\journal{arXiv}

\begin{document}

\begin{frontmatter}

\title{A Comparison of Public Causal Search Packages on Linear, Gaussian Data with No Latent Variables\footnote{Revision 1}}

\author{Joseph D. Ramsey and Bryan Andrews}

\address{Pittsburgh, PA, USA}

\begin{abstract}
We compare Tetrad (Java) algorithms to the other public software packages BNT (Bayes Net Toolbox, Matlab), pcalg (R), bnlearn (R) on the ``vanilla'' task of recovering DAG structure to the extent possible from data generated recursively from linear, Gaussian structure equation models (SEMs) with no latent variables, for random graphs, with no additional knowledge of variable order or adjacency structure, and without additional specification of intervention information. Each one of the above packages offers at least one implementation suitable to this purpose. We compare them on adjacency and orientation accuracy as well as time performance, for fixed datasets. We vary the number of variables, the number of samples, and the density of graph, for a total of 27 combinations, averaging all statistics over 10 runs, for a total of 270 datasets. All runs are carried out on the same machine and on their native platforms. An interactive visualization tool is provided for the reader who wishes to know more than can be documented explicitly in this report.
\end{abstract}


\end{frontmatter}

\section{Introduction}

Understanding how variables relate to one another causally has been of interest in many fields, including biology, neuroscience, climate science, economics, and the social sciences. It has been argued that causal research is best carried through intervention of variables or through controlled randomized experiments, though this is often prohibitively expensive, unethical, or impossible. An alternative is the discovery of causal relations from observational data alone. Numerous algorithms have been constructed to this end, which fall generally into one of two groups, constraint-based (like the PC algorithm) or score-based (like the GES algorithm). In addition to understanding how these algorithms relate to one another theoretically, it is at a basic level of interest to see how they related to one another in performance. To this end, we compare Tetrad implementations (in Java)\footnote{http://www.phil.cmu.edu/tetrad/} to a number of widely-used implementations from publicly accessible packages: BNT (Matlab), pcalg (R, C),\footnote{https://cran.r-project.org/package=pcalg} bnlearn (R, C++),\footnote{http://www.bnlearn.com/} on their native platforms and on the same machine. Notably, through the Center for Causal Discovery\footnote{\url{http://www.ccd.pitt.edu/}}, the Tetrad algorithms have been given wrappers for R and Python, so there is no need usually for researchers to give up their accustomed platform in order to use them. We compare them, however, in their native Java.\footnote{http://www.phil.cmu.edu/tetrad/}

We focus on what we call the ``vanilla'' task--data simulated i.i.d. recursively from linear, Gaussian structural equation models (SEMs), with no latent variables, from graphs that are not particularly ``surprising'', and with no intervention data. To make random graphs, we simply add edge randomly in the forward direction to variables in a linear order. Coefficients for edges are drawn uniformly from (0.2, 0.9); exogenous error variances from (1, 3).\footnote{A ``surprising'' graph might be one that is, for example, scale-free, so that some nodes have many adjacents, others very few. Typically nodes with many adjacents are difficult to analyze unless all arrows point out from them.} We study this task, because each of the platforms studied has at least one algorithm to address it, so the comparisons from one platform to the next on the same datasets are meaningful. We vary sample size as 100, 500 or 1000; we vary number of variables in the generating graph as 50, 100, or 500; we vary the average degree of the graph as 2, 4, or 6. For each combination, we simulate and store 10 random graphs and datasets, with different coefficients and exogenous error variances.

We recognize that there are many types of data on which one may wish to compare causal search algorithms. One may wish to look at linear or nonlinear connection functions, Gaussian or non-Gaussian exogenous errors, continuous or discrete variables, or a mixture continuous and discrete.\footnote{See, for example, Andrews et al., 2017} One may wish to assume knowledge of time order or not, or knowledge of adjacent structure. We take the attitude, though, that these problems need to be taken up one at a time as algorithms become publicly available. Since the ``vanilla'' task has taken up so much of the effort of so many researchers over so many years, it would be good to get a snapshot of progress on it. We will draw some very basic conclusions from this study after presentation of results. Our attitude is simply that, while there is great value in comparing algorithms based on their theoretical virtues, there is also at a base level value in comparing them head to head on a common task, for common parameter values, on a common machine, just on their performance. This way, the strategies that are most effective quickly become apparent and may be pursued further, and strategies that aren't as effective may be pursued, after some thought, perhaps less vigorously.

We run each algorithm on its native platform (Matlab, R, or Java), on exactly the same datasets and on exactly the same machine, and compile the results into a single table. We believe this renders the results comparable. To our knowledge, there is no such previous comparison of all of these implementations together, though comparisons of subsets have been pursued.

The table of results that comes out of this is quite large, too large really to draw good conclusions by eye without a tool to do so. To this end, we offer an interactive visualization comparison of the results. We give some results individually in the Results section below, but the reader is encouraged to try the visualization tool and draw their own conclusions.

In Section 2 we give the simulation task on which we are comparing the various algorithms. In Section 3, we specify the comparison procedure. In Section 4, we preview the packages and algorithms we will compare. In Section 5, we describe the visualization tool we will use. In Section 6, we give some sample results. In Section 7, we give a discussion of these results. In Section 8, we draw some conclusions. It should be noted (as we will in fact note) that there are many more results than are described in the text of the paper; we give these results in an Appendix. This table of results is available as an attached file with this report for use with the visualization tool. See Section 5.

\section{The Simulation Task}

We generate data, at various sample sizes, from models where they struture should be recoverable, given unbounded sample size. Most of the algorithms listed can recover the pattern (CPDAG) of the true DAG, some the adjacency structure of the true DAG.  We are interested to know how the algorithms scale in three dimensions. First, we'd like to know how they scale up with sample size. Second, we'd like to know how they scale up with the number of variables being studied. Third, we'd like to know how they scale up with the density of the generating graph. We, therefore, create datasets for models that are varied along these three dimensions. For sample size, we chose 100, 500, or 1000 samples. 100 we will take to be a ``small'' sample, although for some problems it's large. 1000 is usually a large sample, where one expects near convergence in the typical case. For numbers of variables, we choose 50, 100, or 500 variables. 500 is not an upper limit for all of the algorithms we study, not by any means. But it is already out of reach for some of them, so in the interest of providing a common set of datasets to study across as many algorithms as possible, we limit it thus. Likewise, 50 is by no means necessarily a small number of variables; for some problems, that's in fact quite large, but we are interested in how these algorithms scale, so we pick 50 as a relatively small number of variables to study. For the density of graph, we vary the average degree of the graph, which is the average total number of parents and children each node has in the true DAG. We choose values of 2, 4, and 6. Again, a graph of average degree 6 is not necessarily the densest graph one may wish to analyze. In fact, for some scientific problems, for instance in neuroscience, the density of the true model may be much greater than that. Nevertheless, an average degree of 6 is already out of reach for many algorithm, including many of the algorithms we are comparing, so we limit ourselves thus.

This yields a total of 27 combinations. For each combination, we simulate ten graphs and data sets randomly choosing simulation parameters for each dataset, for a total of 270 datasets. These are stored in a public GitHub repository.\footnote{https://github.com/cmu-phil/comparison-data}; readers are encouraged to try their hand at these datasets to see if they can improve on the performances described here.

\section{The Comparison Procedure}

We adapted the Algorithm Comparison Tool in Tetrad \cite{ramsey2016comparing} to do the comparisons. That tool automates the simulation of datasets and the running of lists of algorithm variants on that data to produce a final tabular report of performance statistics. For various reasons, this was ill-suited to the task of comparing algorithms cross-platform, so adjustments were made. First, elapsed times were in the original tool calculated automatically on the fly for algorithms without a record being made of them; we explicitly made a record of these. Second, result graphs for the existing tool were stored internally and results tabulated over them; as an adjustment we stored these out in a directory structure to files. Third, since graphs on other platforms are saved in formats different from Tetrad's graph format, parsers needed to be written to read graphs from each platform, sometimes for each algorithm, so that in the end all graphs could be compared to the correct ``true'' DAG for each simulation; we wrote these parsers. Graph outputs and elapsed time files were stored in parallel directories and classes were added to reload the original simulation data and true DAGs, the result graphs and elapsed times, and to calculate performance statistics and present them in a table in the manner of the original algorithm comparison tool. 

Importantly, all algorithms were run on their native platforms, with timing results calculated on those platforms. To make the timing results comparable, we ran all algorithms on the same machine, with 3.3 GHz Intel I7 processor, with 16G RAM. This processor has two cores with two hardware threads each, so there is some speedup available for algorithms that can take advantage of parallelization. The version of Matlab used was 9.2.0.556344 (R2017a); the version of R used was 3.4.1; the version of Java used was 1.8.0\_131. The version of BNT used was last updated 17 Oct 2007; a newer version could not be found. The pcalg code used was version 2.5.0-0, packaged 2017-07-11 and built from sources. The version of bnlearn used was 4.2, published 2017-07-03, and built from sources. The version of Tetrad used was commit 39b8abd in the Tetrad GitHub repository from 8/3/2017. The end result of this analysis was a table of simulation results for each algorithm or algorithm variant, each sample size, each number of variables, and each average degree. All statistics were averaged over 10 runs.

While the algorithms comparison tool allows arbitrary statistics to be calculated and recorded in the final table, we settle on the following list of statistics, which seem to us to be fairly expressive. Let $ATP$ be the number of true positive adjacency judgments, $AFP$ the number of false positive adjacency judgments, $AFN$ the number of false negative adjacency judgments; let $AHTP$, $AHFP$, $AHFN$ be the corresponding judgments for arrowheads in the graph, where $X \rightarrow Y$ counts as as true positive arrowhead if it occurs in both the true DAG and the estimated graph. Then we have the following measures:

Vars: Number of variables in the graph

Deg: Average degree of the graph

N: Number of simulated samples

$AP$: Adjacency Precision = $\frac{ATP}{ATP + AFP}$

$AR$: Adjacency Recall = $\frac{ATP}{ATP + AFN}$

$AHP$: Arrowhead precision = $\frac{AHTP}{AHTP + AHFP}$

$AHR$: Arrowhead recall = $\frac{AHTP}{AHTP + AHFN}$

$McAdj$: Matthew's correlation coefficient for adjacencies
\begin{center}
$= \frac{(ATP * ATN) - (AFP * AFN)}{\sqrt{(AFP + AFP) *(ATP + AFN) * (ATN + AFP) * (ATN + AFN)}}$
\end{center}

$McArrow$: Matthew's correlation coefficient for arrowheads
\begin{center}
$= \frac{(AHTP * AHTN) - (AHFP * AHFN)}
{\sqrt{(AHFP + AHFP) *(AHTP + AHFN) * (AHTN + AHFP) * (AHTN + AHFN)}}$
\end{center}

$E$ = Elapsed Time in Seconds

Here, the Matthews correlation ranges from -1 to 1, with higher positive values indicating greater agreement between precision and recall, respectively, for adjacency or arrowhead judgments. The Matthews correlation is designed to give an aggregate accuracy for precision and recall, balancing well when the sizes of the constituent sets (TP, FP, FN) are very different in size. We report $AP$, $AR$, $AHP$, $AHR$, and $E$ in the results section below but include $McAdj$ and $McArrow$ in the tables in the Appendix.

In some cases results are not provided. These are cases where running time for a single run of the algorithm was greater than 10 minutes.

\section{The Packages and their Algorithms Relevant to this Task}

We make some general remarks that apply to multiple packages. Algorithms in these packages, as is generally the case, are commonly said to be of two sorts, \emph{constraint-based} or \emph{score-based}. The distinction is blurred, as has been noted by \cite{nandy2015high}. 

The PC-style algorithms are all constraint-based because they operate entirely from considerations of conditional independence, which constitute constraints on the search. They take a conditional independence test as an \emph{oracle}; the data themselves are screened off by the conditional independence test. In fact, the algorithms could operate without data if the conditional independence facts alone were available. So a sensible way to check the correctness of these algorithms is to feed them d-separation facts from a known true graph and see if they can recover the CPDAG or adjacencies of the DAG, depending on the algorithm. 

The GES algorithm (Chickering 2002), by contrast, is said to be score-based, on the strength of the assertion that they rely on Bayesian scores like BIC for their search. However, it can also be carried using conditional independence tests as oracles. To make this point, we include some runs of FGES for Tetrad using the Fisher Z test. For the Fisher Z runs, we use $\alpha - p$ as a ``score'', where alpha is the alpha cutoff for PC and $p$ is the p-value for the conditional independence test. This is not a proper score, but it does convey the conditional independence information properly, and as can be seen, it works well. Also, we have undertaken to check the correctness of FGES from d-separation facts alone. One way to do this is to return a score of +1 for conditional d-connection and -1 for conditional d-separation. If one does this, GES should, in fact, give the same graphs back from d-separation facts alone as PC, and it does, for all of the FGES implementations. We assume similar checks have been made for pcalg. In general, however, using a score for GES that prioritizes true colliders changes the priority of addition and removal of edges in the forward and backward stages, in ways that help to find true colliders more quickly. This shortens and stabilizes the search.

For the PC-style algorithms, for small samples, there is always the possibility of a collider conflict--that is, a situation where one is given a model $X \rightarrow Y \leftarrow Z - W$, where the $X - Y - Z$  triple has been oriented as a collider, and for similar reasons one wants to orient the triple $Y - Z - W$ as a collider as well. For the PC algorithm this can happen if the edge $X - Z$ has been removed from the graph on the strength of the observation that $I(X, Z | S1)$ where Y is not in S1, but also $I(Y, W | S2)$ where X is not in S2. (Here, $I(X, Y | Z1)$ is true if and only if X is independent of Y conditional on set Z of variables.) For models in which there are no latent variables, this is a small sample error, but it may occur quite often, depending on the algorithm. There are at least three possible graphs that could result. One could orient both colliders to get $X \rightarrow Y \leftrightarrow Z \leftarrow W$; we call this the \emph{bidirected rule}; this has been pursued in the Tetrad package in previous versions.\footnote{The reason this has been pursued is that when latent common causes of pairs of measured variables affect the data, a bidirected edge of this form indicates the existence of one or more latent variables, a desired behavior. Here, since we know there are no latent variables, these bidirected edges are a nuisance, due to small sample errors of judgment, so we don't try to orient them.} Or one could orient just the first collider and refuse to orient the second, which we call the \emph{priority rule}--that is, $X \rightarrow Y \leftarrow Z - W$. Or one could simply orient the second without unorienting the $X \rightarrow Y$ edge, to get $X \rightarrow Y \rightarrow Z \leftarrow W$, which we call the \emph{overwrite rule}. For these runs, Tetrad's PC algorithms are all using the priority rule. The pcalg package below uses the overwrite rule.

For constraint-based algorithms we use the default conditional independence test for the package, except where noted. When there is an option for setting an alpha level, we set it at at 0.01 or 0.001, except where noted. For the GES algorithms, we use BIC, with a multiplier on the penalty, so $BIC = 2L - c k ln N$, where $L$ is the likelihood, $c$ the penalty discount, $k$ the degrees of freedom, and $N$ the sample size. We use penalty discounts 2 and 4; for pcalg, this corresponds to $
\lambda$'s of $2 ln N$ and $4 ln N$ (see \cite{nandy2015high}.

Scripts used to generate all results are available. For Tetrad, they are included in the GitHub results repository\footnote{https://github.com/cmu-phil/causal-comparisons; this repository is currently private because other other incomplete comparisons are included, but will be made available to readers if a GitHub ID is sent to the authors. But all of the results graphs and scripts are included there.} in a directory called ``condition2''.

Since we were adapting the Tetrad algorithms comparison tool, we began with the Tetrad algorithms and proceeded from there.

\subsection{The Tetrad Suite}

The Tetrad package contains more or less the same variety of algorithms for the ``vanilla'' case as pcalg, with some notable variations in particular cases. It contains almost all of the same variation in PC-style algorithms (PC, PC-Stable, CPC, CPC-Stable) and also an optimized GES algorithm (FGES). The package is written in Java. The indices given are the indices of the variants in the Appendix.

The PC algorithm was defined in \cite{spirtes1993search}; \cite{meek1995causal} later gave implied orientation rules which have been adapted in the Tetrad implementation. The PC-Stable algorithm \cite{colombo2014order} aims to render the adjacencies in the output independent of the order of the input variables; we have implemented this based on the description. The PC-Stable-Max algorithm alters the CPC algorithm to orient colliders using maximum p-value conditioning sets and renders both adjacencies and orientations order-independent. The CPC algorithm (\cite{ramsey2012adjacency}) orients colliders by insisting on uniformity of independence judgments. CPC-Stable combines CPC orientation with the PC-Stable adjacency procedure. The ``stable'' adjacency search (\cite{colombo2014order}) was adapted from the pcalg package.

Notably, for the CPC algorithm, we interpret all ambiguous markings in the graph as noncolliders (see \cite{ramsey2012adjacency}). CPC aims to mark all unshielded triple $X - Y - Z$ as unambiguous colliders, unambiguous noncolliders, or ambiguous. Thus the output represents a \emph{collections of CPDAGs}, not a specific CPDAG. To get a specific CPDAG from the collections, one needs to choose for each ambiguous triple whether it is a collider or a noncollider, and then apply the Meek orientation rules. Here, we choose to interpret each ambiguous triple as a noncollider, as does pcalg. We do this for compatibility.

The FGES algorithm (\cite{ramsey2017million}) modifies the GES algorithm (\cite{chickering2002optimal}), with the intention of making it more scalable. It forgoes the full loop in the forward phase, using caching  of information instead to avoid some repetition of tests. There are two settings, faithfulness = true and faithfulness = false. If the faithfulness flag is set to true, edges $X - Y$ with $X$ uncorrelated with $Y$ are not added to the graph in a first sweep; for paths $X - Y - Z$ that should be colliders, $X - Z$ is then added to the graph and removed in a subsequent backward step, orienting the collider. This strategy is similar to the one described in \cite{nandy2015high} for the ARGES algorithm and results in significant speedup, as they attest. With the faithfulness flag set to false, the same first sweep (forward followed by backward) is performed, but then edges $W - Z$ are recovered for $X$ that are unconditionally d-connected to $Z$ but not correlated with $Z$ unconditionally in a second forward sweep, and the backward sweep is repeated. This allows for some edges with canceling paths to be recovered and is equivalent to Chickering's formulation of GES. Additionally, sections of the algorithm that can be parallelized are treated as parallelized. This includes in particular the initial step of adding a single edge to an empty graph, a step which is particularly time-consuming.

As noted, all g of the PC variants for the Tetrad runs are configured to use the \emph{priority rule} to resolve colliders. An alpha of 0.01 and 0.001 are used. For FGES, some smaller, for Fisher Z, some smaller alphas are included, 0.0001 and 0.00000001, since these are advantageous. For FGES with BIC, penalties 2 and 4 are included. The following are the specific variants include. The indices correspond to the indices in the full in the Appendix.

1. PC (``Peter and Clark''), Priority Rule, using Fisher Z test, alpha = 0.01

2. PC (``Peter and Clark''), Priority Rule, using Fisher Z test, alpha = 0.001

3. PC-Stable (``Peter and Clark'' Stable), Priority Rule, using Fisher Z test, alpha = 0.01

4. PC-Stable (``Peter and Clark'' Stable), Priority Rule, using Fisher Z test, alpha = 0.001

5. PC-Stable-Max (``Peter and Clark''), Priority Rule, using Fisher Z test, alpha = 0.01

6. PC-Stable-Max (``Peter and Clark''), Priority Rule, using Fisher Z test, alpha = 0.001

7. CPC (Conservative ``Peter and Clark''), Priority Rule, using Fisher Z test, alpha = 0.01

8. CPC (Conservative ``Peter and Clark''), Priority Rule, using Fisher Z test, alpha = 0.001

9. CPC-Stable (Conservative ``Peter and Clark'' Stable), Priority Rule, using Fisher Z test, alpha = 0.01

10. CPC-Stable (Conservative ``Peter and Clark'' Stable), Priority Rule, using Fisher Z test, alpha = 0.001

11. FGES (Fast Greedy Equivalence Search) using Fisher Z Score, alpha = 0.001, faithfulnessAssumed = false

12. FGES (Fast Greedy Equivalence Search) using Fisher Z Score, alpha = 1.0E-4, faithfulnessAssumed = false

13. FGES (Fast Greedy Equivalence Search) using Fisher Z Score, alpha = 1.0E-8, faithfulnessAssumed = false

14. FGES (Fast Greedy Equivalence Search) using Sem BIC Score, penaltyDiscount = 2, faithfulnessAssumed = false

15. FGES (Fast Greedy Equivalence Search) using Sem BIC Score, penaltyDiscount = 4, faithfulnessAssumed = false

16. FGES (Fast Greedy Equivalence Search) using Fisher Z Score, alpha = 0.001, faithfulnessAssumed = true

17. FGES (Fast Greedy Equivalence Search) using Fisher Z Score, alpha = 1.0E-4, faithfulnessAssumed = true

18. FGES (Fast Greedy Equivalence Search) using Fisher Z Score, alpha = 1.0E-8, faithfulnessAssumed = true

19. FGES (Fast Greedy Equivalence Search) using Sem BIC Score, penaltyDiscount = 2, faithfulnessAssumed = true

20. FGES (Fast Greedy Equivalence Search) using Sem BIC Score, penaltyDiscount = 4, faithfulnessAssumed = true

\subsection{The bnlearn Package}

The blearn package (\cite{tsamardinos2003time}, \cite{tsamardinos2003algorithms}, \cite{tsamardinos2006max}, \cite{aliferis2010local}) consists of a variety of algorithms, largely centered around the idea of finding Markov blankets and using Markov blankets to construct estimations of CPDAGs or undirected graph representing the adjacency structure of the true DAG. Judging from their description, these algorithms are not primarily aimed at the purely continuous case; we will come back to them in a subsequent study of multinomial data, for which more complete test results from the authors are available. Nevertheless, we give the continuous-only results here, using the default conditional independence test.

Two of the algorithms, MMPC and HITON, estimate adjacencies but not orientations. For this reason, orientation statistics will not be given for them.

The bnlearn algorithms are all written natively in C and included in R through the usual sort of wrappers. 

We test the following algorithms and their variants. (the indices are the indices of these variants in our output table.) We use an alpha of 0.01 or 0.001. MMHC in the version of the software we tested does not take an alpha values.

21. MMPC alpha = 0.01

22. MMPC alpha = 0.001

23. GrowShrink alpha = 0.01

24. GrowShrink alpha = 0.001

25. IAMB alpha = 0.01

26. IAMB alpha = 0.001

27. Fast.IAMB alpha = 0.01

28. Fast.IAMB alpha = 0.001

29. Inter.IAMB alpha = 0.01

30. Inter.IAMB alpha = 0.001

31. si.hiton.pc alpha = 0.01

32. si.hiton.pc alpha = 0.001

33. MMHC

Finally, we took one of the bnlearn algorithms, iamb, and varied the independence test from the available options. The ``monte carlo'' (mc) tests did not scale well even to our smallest problems, so we focused on the options without the Monte Carlo adjustment.

34. iamb alpha = 0.01.test=cor

35. iamb alpha = 0.001.test=cor

36. iamb alpha = 0.01.test=mi-g

37. iamb alpha = 0.001.test=mi-g

38. iamb alpha = 0.01.test=mi-g-sh

39. iamb alpha = 0.001.test=mi-g-sh

40. iamb alpha = 0.01.test=zf

41. iamb alpha = 0.001.test=zf

\subsection{The pcalg Package}

Descriptions for the pcalg package may be found in \cite{kalisch2012causal}; for more recent documentation, please see the documentation provided in the R package pcalg. We focus just one algorithms that apply directly to the ``vanilla'' task: the PC-style algorithms and the GES implementation. As noted above, pcalg and Tetrad for this case contain very similar implementation. They differ in implementation style, language, and platform. The pcalg algorithms are all written natively in C++ and included in R through the usual sort of wrappers. Exploring differences in implementation style and choices is an interesting topic, but this lies beyond the scope of this paper; needless to say, all of the code is publicly available for anyone who is interested.

In pcalg, the $pc$ method is designed to adjust its behavior in response to parameters to morph into the various PC-style algorithms. Most of these have counterparts in the Tetrad suite; some do not (like the \emph{majority rule}). To make the comparison easy, we will use the Tetrad names for the algorithms. The specific pcalg commands used to run those algorithms are available in the \emph{comparison2} directory of the results repository. In all cases, we use alpha values of 0.01 or 0.001. For PC-stable, which is readily parallelizable, we set the number of cores to 4.

The GES method in the pcalg package uses a $\lambda$ rather than a penalty discount parameter, but these are inter-convertible, we use $\lambda = c ln N$; here, $c$ is the penalty discount.

42. PC pcalg defaults alpha = 0.01

43. PC pcalg defaults alpha = 0.001

44. PC-Stable pcalg ncores=4 alpha = 0.01

45. PC-Stable pcalg ncores=4 alpha = 0.001

46. CPC pcalg defaults alpha = 0.01

47. CPC pcalg defaults alpha = 0.001

48. CPC pcalg majority.rule defaults alpha = 0.01

49. CPC pcalg majority.rule defaults alpha = 0.001

For GES, we test the following variants:

50. GES pcalg defaults 2*log(nrow(data))

51. GES pcalg defaults 4*log(nrow(data))

\subsection{The BNT Package}

BNT (``Bayes Net Toolbox'', \cite{murphy2001bayes}) contains a number of algorithms, Once one eliminates algorithms that are for discrete variables only, algorithms that assume time order of the variables or knowledge of adjacency structure, or algorithms that are designed for the latent variable case, the only structure learning algorithm left to compare with is the implementation of the PC algorithm, learn\_struct\_dag\_pc. This takes an alpha level; we use 0.01 or 0.001. The BNT algorithms are written natively in Matlab. There are two variants for PC in BNT in our table:

52. learn.struct.pdag.pc bnt alpha = 0.01

53. learn.struct.pdag.pc bnt alpha = 0.001

\section{Visualization Tool}

The results are shown in the Appendix in tabular form. Since the this table is quite large, we found it useful to provide an interactive visualization tool\footnote{\url{http://www.pitt.edu/~bja43/causal}} to help readers understand the results. We render figures from this tool in the Results section below. However, we invite interested readers to explore it on their own, as there is much more data to explore (see Appendix).  One simply needs to download these three files, stats.txt and config.txt, and std.txt,\footnote{These files are currently provided as example files for the visualization tool, but they are also included in this arXiv download.} then go to the URL above for the tool and load them the respective buttons, then set the number of runs to 10. By providing a file of standard deviations (std.txt) and the number of runs, the tool additionally calculates 95\% confidence intervals around the mean statistics; \cite{YujiSODE2017} is used to calculate the inverse $t$ CDF.  Then click to the Plot tab and select a combination of parameters for which a result exists, and bar plots will be rendered as in the Results section, below. As many or as few results as are desired may be simultaneously plotted. 

In order for the bar plots to be displayed, one needs to select a combination of parameters for which data is available in the table. One needs to pick a one or more numbers of variables, average degrees, and sample sizes. One needs to pick parameters for which they are compared. (Different algorithms may require different parameters.) Then one needs to pick one or more algorithms from the four packages.\footnote{It is worth noting that this visualization tool is intended to be useful outside of this particular project, and already has been. One simply needs to make a stats, config, and std files in the proper format and load them in. Plots may be edited using Plotly.} Provided one has selected a combination of settings for which data is available, charts will be rendered.

\section{Results}

\begin{figure}
  \centering
    \includegraphics[width=\textwidth]{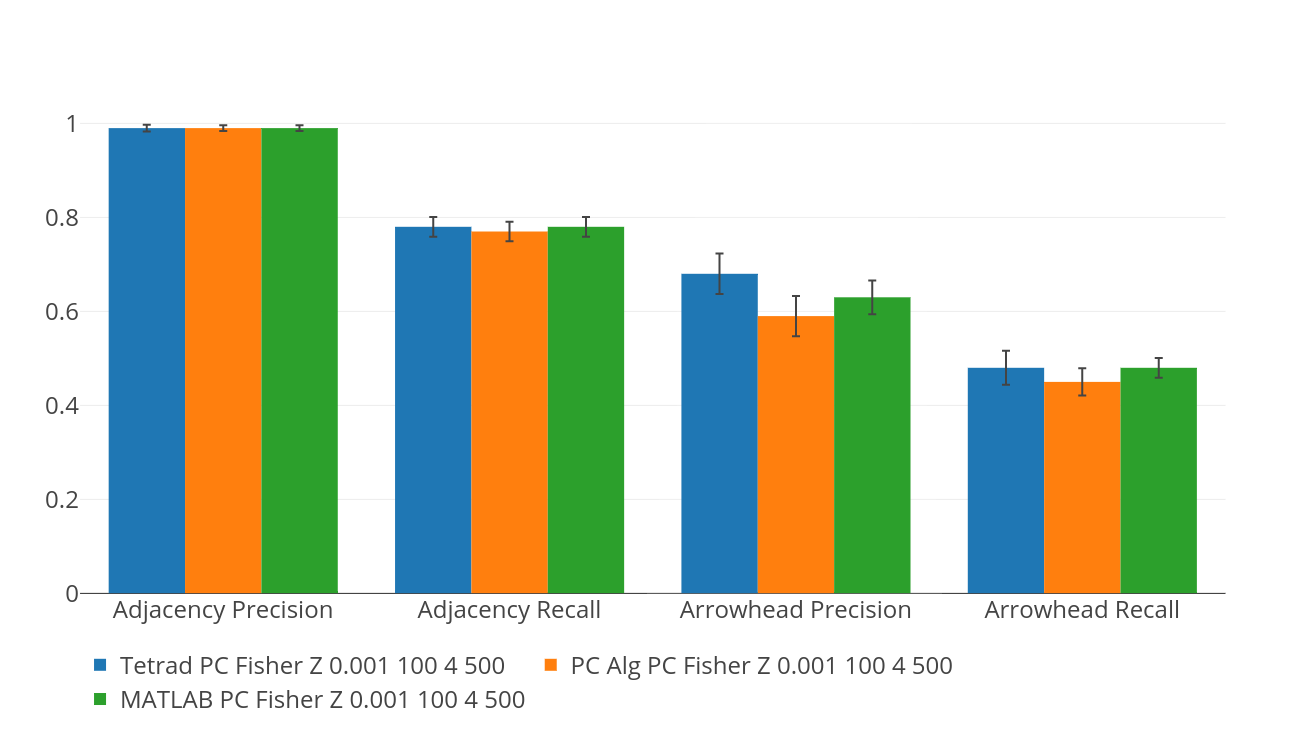}
  \caption{Adjacency and orientation accuracy of PC across platforms for 100 variables, average degree 4, 500 samples, alpha = 0.001.}
  \label{pc_3packages}
\end{figure}

There is one algorithm that is implemented on three different of the platforms--viz., PC. We show results for this in Figure \ref{pc_3packages}. Here we use an alpha value of 0.001, with 100 variables, average degree 4, sample size 500.

\begin{figure}
  \centering
    \includegraphics[width=\textwidth]{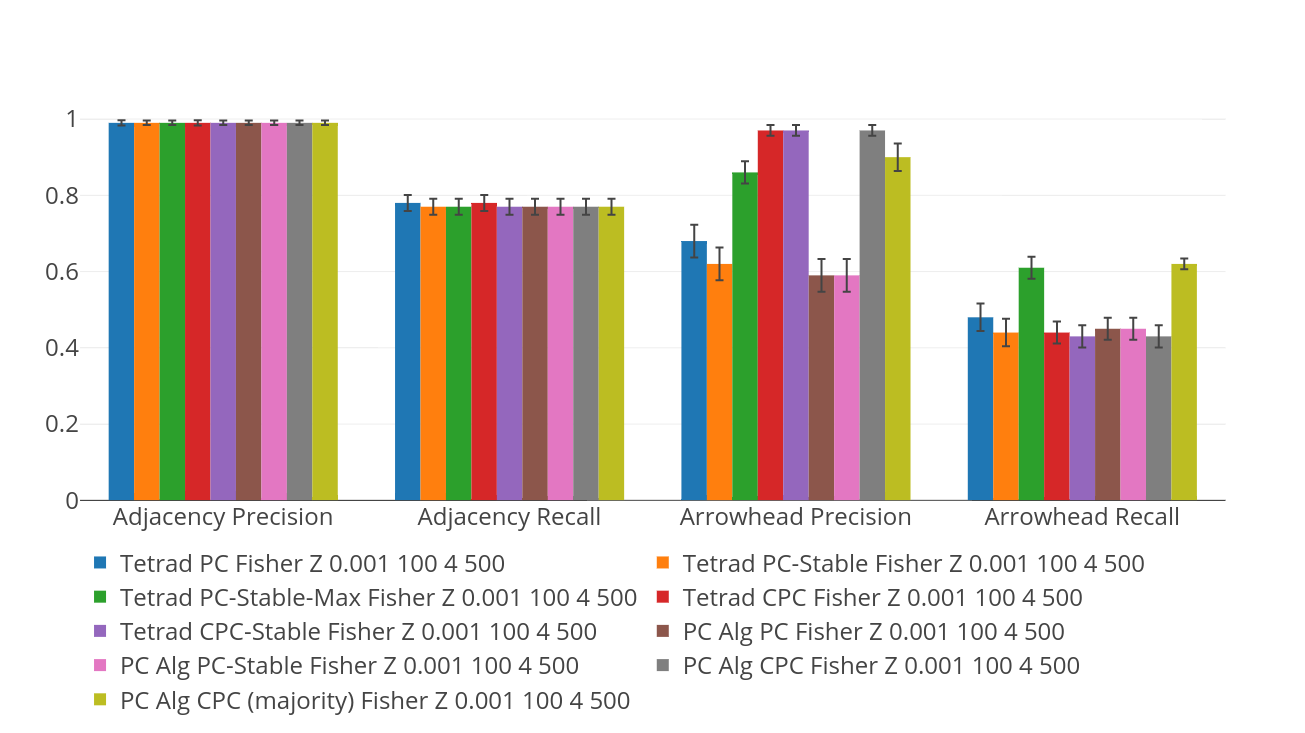}
  \caption{Adjacency and orientation accuracy of PC-style algorithm for Tetrad and pcalg platforms. This is for 100 variables, average degree 4, sample size 500, alpha = 0.001.}
  \label{tetrad_pcalg_pc_algorithms}
\end{figure}

In Figure \ref{tetrad_pcalg_pc_algorithms} we show a comparison of all of the PC-style algorithms in the Tetrad and pcalg packages. Here, we look at the 100 variables case, with average degree 4, sample size 500, alpha = 0.001.

\begin{figure}
  \centering
    \includegraphics[width=\textwidth]{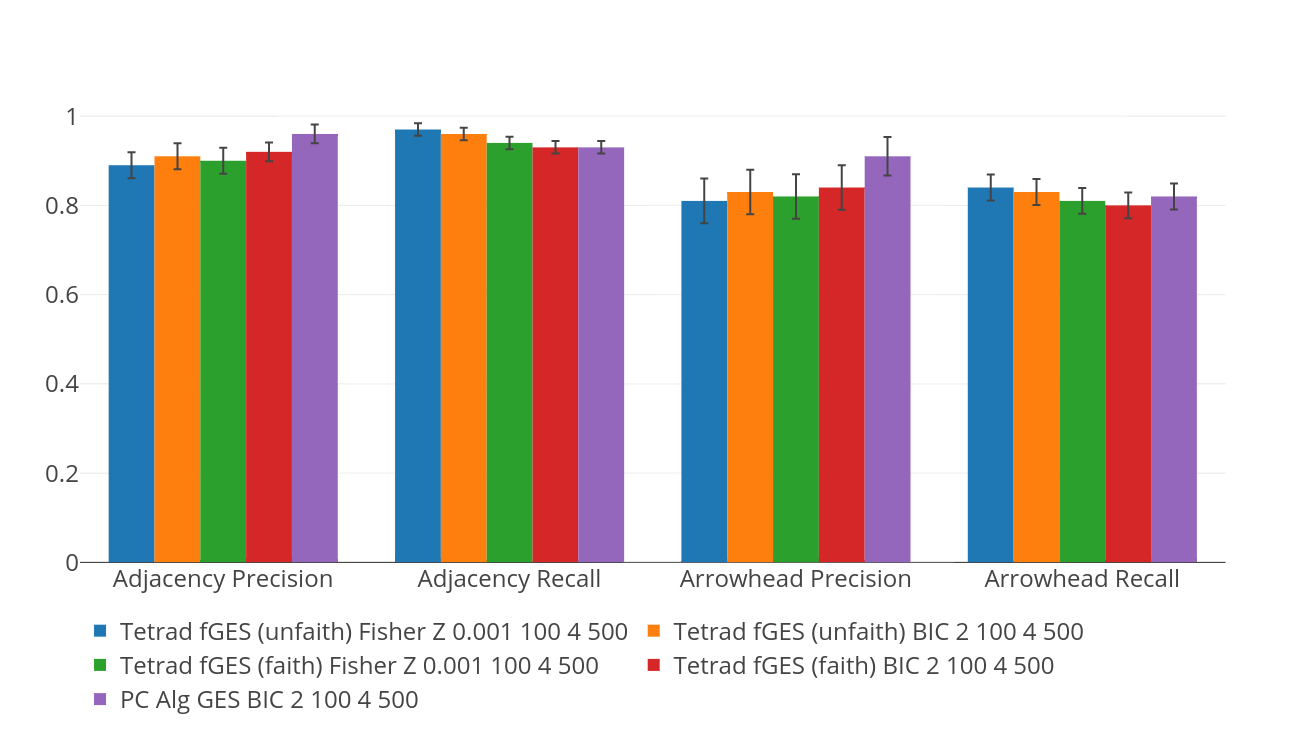}
  \caption{Adjacency and orientation accuracy of GES for Tetrad and pcalg packages. This is for 100 variables, average degree 4, sample size 500. For Fisher Z algorithms, alpha = 0.001; for BIC algorithms, penalty = 2 (for pcalg, $\lambda = 2 ln N$).}
  \label{tetrad_pcalg_ges_algorithms}
\end{figure}

In Figure \ref{tetrad_pcalg_ges_algorithms} we show a comparison of all of the GES algorithms in the Tetrad and pcalg packages. Here, we look at the 100 variable case, average degree 4, sample size 500. For algorithms that use the Fisher Z test, an alpha of 0.001 was used; for algorithms that use a BIC score, a penalty of 2 was used. (For pcalg, $\lambda$ in this case is $2 ln N$.) 

\begin{figure}
  \centering
    \includegraphics[width=\textwidth]{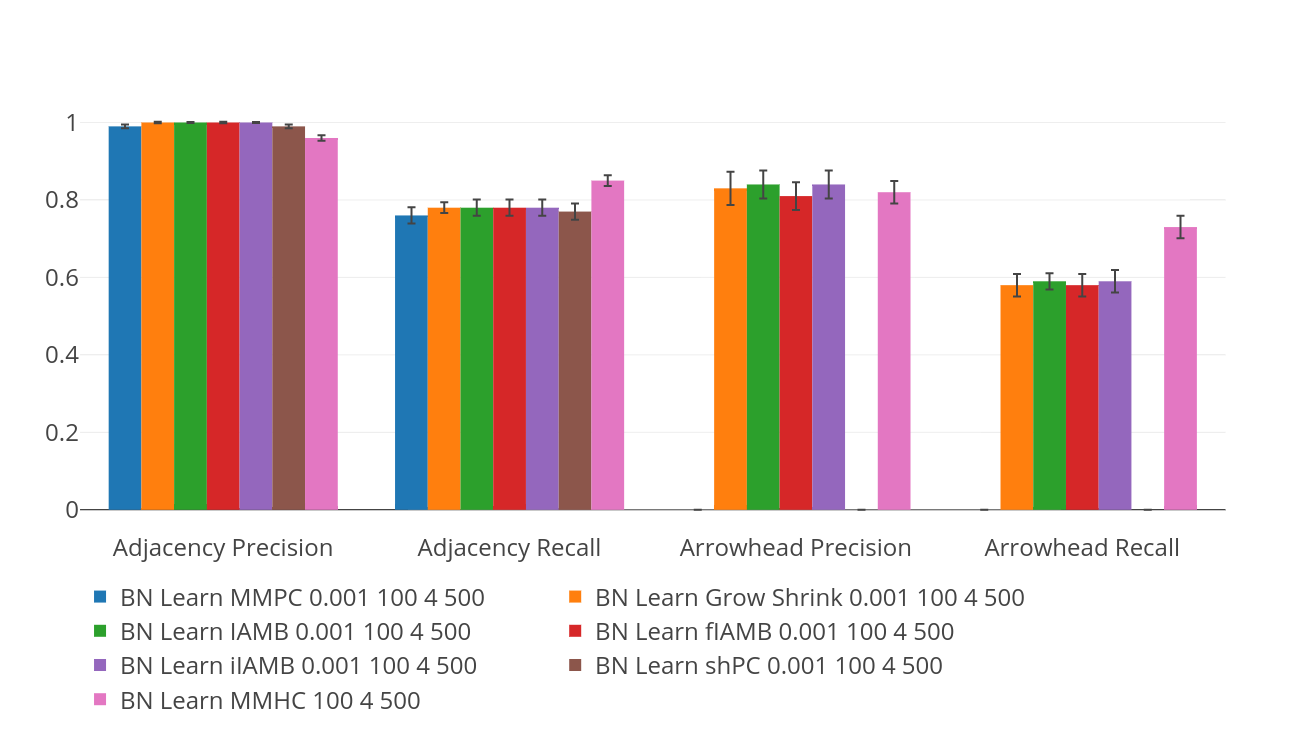}
  \caption{Adjacency and orientation accuracy of bnlearn algorithms. Algorithms with no orientation statistics do not estimate orientations. This is for 100 variables, average degree 4, 500 samples, alpha = 0.001, where applicable.}
  \label{bnlearn_algorithms}
\end{figure}

In Figure \ref{bnlearn_algorithms} we show a comparison of all of the algorithms tested in the bnlearn package. Here, we look at the 100 variable case, average degree 4, sample size 500, alpha = 0.001. 

The plots in Figure \ref{pc_3packages}, Figure \ref{tetrad_pcalg_pc_algorithms}, \ref{tetrad_pcalg_ges_algorithms}, and \ref{bnlearn_algorithms} are all for the same datasets and so are comparable. Using the visualization tool one may obtain similar plots for other cases.

\begin{figure}
  \centering
    \includegraphics[width=\textwidth]{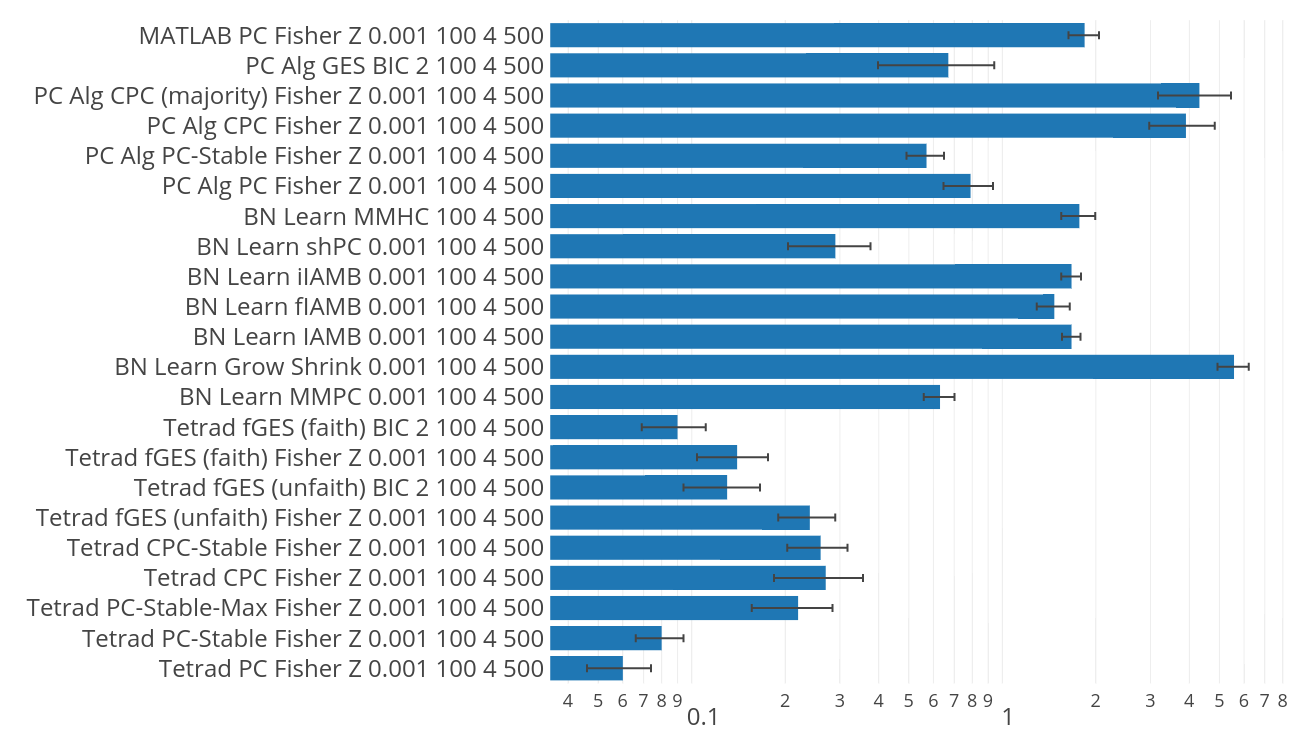}
  \caption{Elapsed times for all algorithms, average degree 4, number of variables 100, sample size 500, alpha 0.001, penalty 2. The elapsed times are on a log scale.}
  \label{all_times}
\end{figure}

In Figure \ref{all_times} we show a comparison of elapsed times for all of the above choices, for common datasets.

\section{Discussion}

For Figure \ref{pc_3packages}, it is important to note that all datasets have had their columns randomly rearranged, so that there is no opportunity to gain an advantage from having the columns in causal order. Under these conditions, PC (as with most algorithms for coefficients in the range +/-(0.2, 0.9)) gets nearly 100\% precision for adjacencies, with recall about 0.8, and orientation precision of about 0.6, with orientation recall (judged against the true DAG) about 0.45. This is a fairly common profile for PC for this sized problem, and all three of the platforms more or less agree to it. The Tetrad algorithms gets slightly higher precision for orientations than the others, perhaps enough to choose it over pcalg, but really the implementations are all very similar.

Figure \ref{tetrad_pcalg_pc_algorithms} compares all of the PC-style algorithms from the pcalg and Tetrad packages. All of these algorithms use the same adjacency search, so it is not surprising that their adjacency statistics are identical. They differ only in orientation. Here, the PC and PC-Stable algorithms yield fairly unusable orientations; the algorithms that attempt to boost collider orientation accuracy all do better. The algorithms that do the best are the CPC variants, for both platforms. There is a trade-off of orientation precision for orientation recall for these algorithms. PC-Stable Max and CPC Majority get the higher orientations recalls, but both suffer somewhat on precision; the performance of each is basically the same.

Figure \ref{tetrad_pcalg_ges_algorithms} compares all of the GES variants from pcalg and Tetrad. Overall, the performances are very similar, though for orientation precision in particular, the pcalg implementation is to be preferred on these grounds.

Figure \ref{bnlearn_algorithms} gives timing results for all of the algorithms compared in the previous figures. The elapsed times are given on a log scale, as some of the times are fairly slow. Generally, the Tetrad algorithms were faster. If we consider CPC to be the most accurate of the PC-style algorithms, for pcalg, it comes back in 3.9 seconds, whereas for Tetrad it comes back in .27 seconds. For GES, the time posted for pcalg was 0.67 seconds, whereas for Tetrad the best time was 0.08 seconds. For this size problem, of course, all of these times are entirely reasonable; one should not be hesitant to wait a few seconds to get a good result.

\begin{figure}
  \centering
    \includegraphics[width=\textwidth]{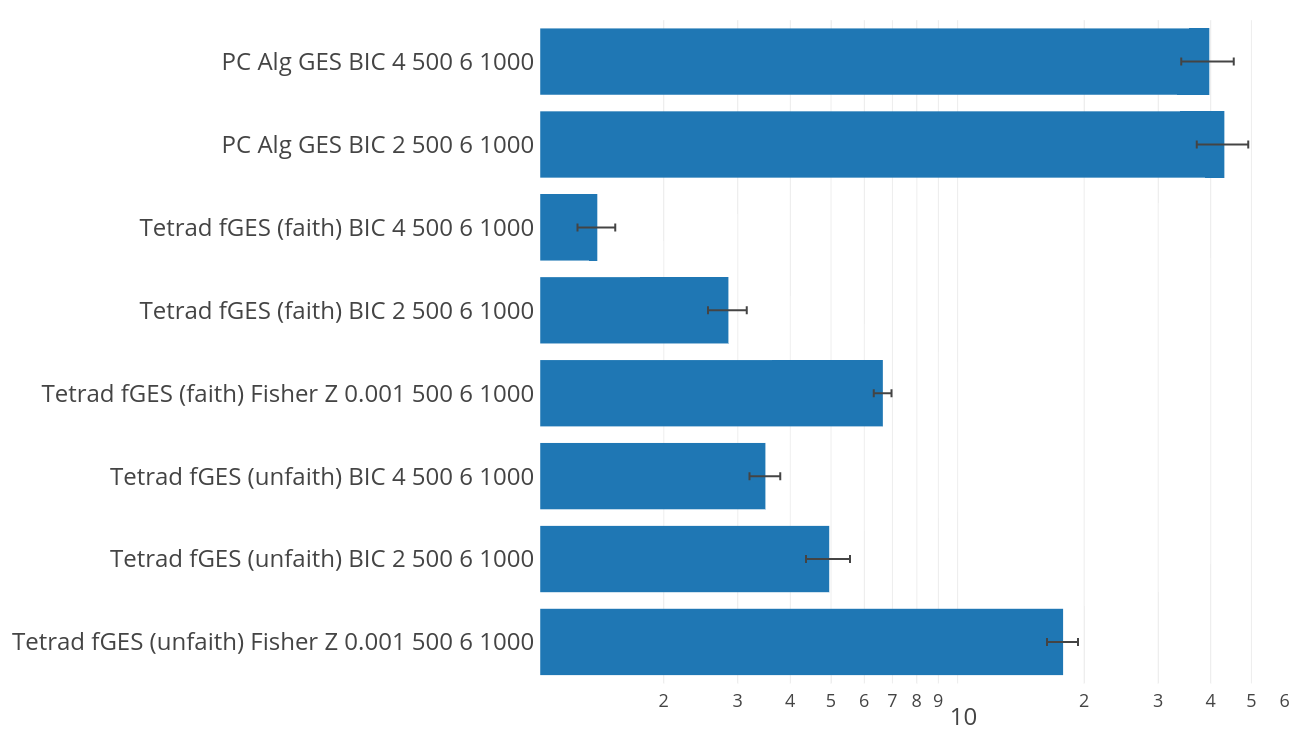}
  \caption{Elapsed times for faster algorithms for the 500 variables, average degree 6, sample size 1000 case, alpha = 0.001, penalty 4. The elapsed times are on a log scale}
  \label{faster_times}
\end{figure}

\begin{figure}
  \centering
    \includegraphics[width=\textwidth]{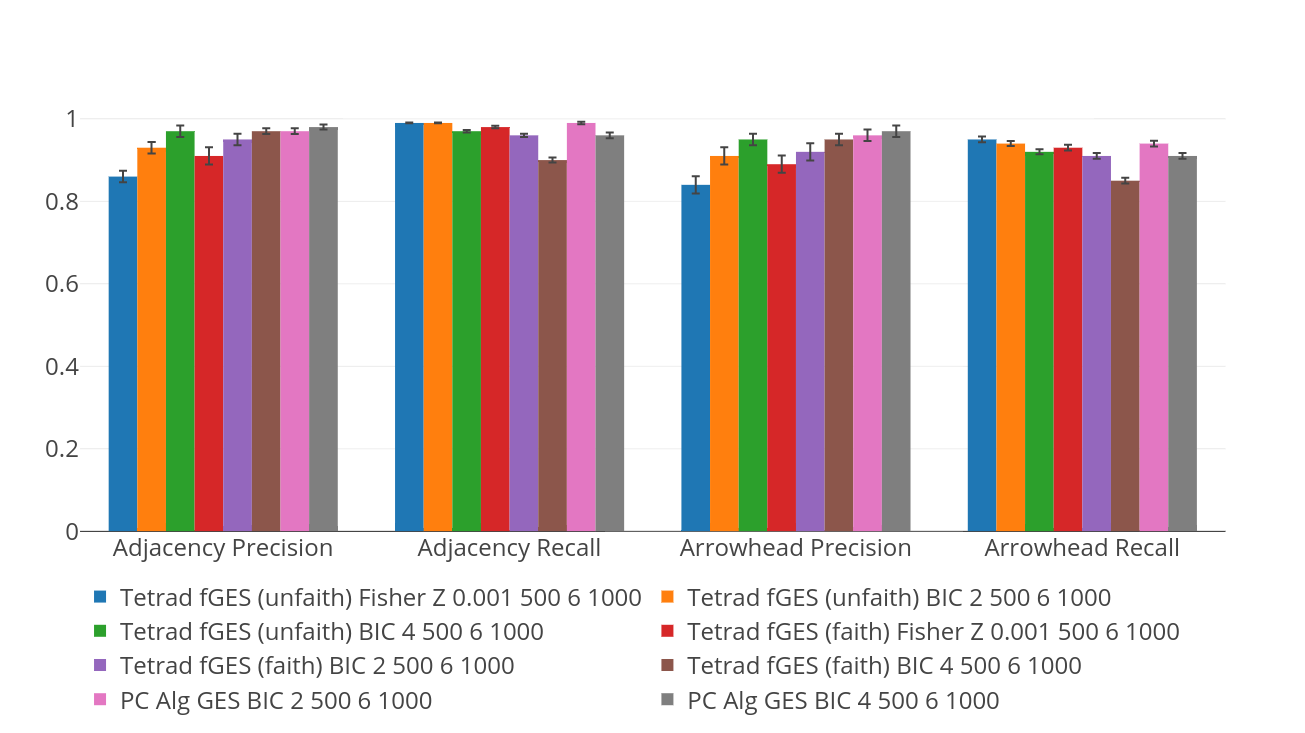}
  \caption{Accuracies for faster algorithms for the 500 variables, average degree 6, sample size 1000 case. The elapsed times are on a log scale.}
  \label{faster_accuracies}
\end{figure}

For scaling up, however, timing may become an issue. In figure \ref{faster_times} we show the faster algorithms, for the largest models we consider, 500 variables, average degree 6, 1000 samples. These are all variants of the FGES algorithms, with one contrast, the pcalg GES algorithm. The fastest of these is for FGES with the faithfulness flag turned on, using the BIC score with penalty 4. In Figure \ref{faster_accuracies} the accuracies of the same algorithms are given. As can be seen, this particular algorithm is not only the fastest at this task but is also one of the most accurate, though the pcalg GES still has a slight advantage in accuracy.

\section{Conclusion}

We have taken an orthogonal approach to comparing algorithms, not based on their theory but based purely on their performance. While we have focused only on the ``vanilla'' task (data generated recursively and i.i.d. from a linear, Gaussian model with no latent variables), this is a task that has been taken up in several software packages, both public like these, but also in private, and which has a long history in the literature. It is usually the first sort of algorithm one tries for a dataset, for better or for worse. We know of no comparison to date that compares all of the public packages to one another directly on this task using exactly the same data on exactly the same machine.

Overall, we find that the Tetrad package fares quite well in this comparison. The PC-style algorithms are comparable to those in other packages. The best of these in terms of orientation precision, CPC, has a similar profile to CPC in pcalg. The main difference is that the Tetrad PC-style algorithms are much faster than their counterparts. There are no doubt algorithmic reasons for this. The pcalg algorithms are implemented in C++, the Tetrad algorithms in Java; generally from benchmark testing C++ is 2 - 2.5 times faster than Java with the same implementation on the same machine. We surmise that if the Java implementation were simply translated intelligently into C++, one would be at least a 2-fold speedup over Java, in this way. Instead the C++ implementations are many times slower. For GES, we find that overall the pcalg implementation has a small but persistent edge over FGES for accuracy, though once again the Tetrad implementation is many times faster. Partly this is due to the parallelization, though the particular machine these algorithms were run on has only two cores, so this doesn't account for the 10 to 20 fold speedup we observe. If one takes into account that C++ is at least two times faster than Java, one might expect that if the Tetrad implementation were directly translated into C++ there would be at least a 20 to 40 fold speedup over the pcalg implementation. This of course suggests a programming project for someone. In any case, this kind of speedup is important, since it outpaces by several fold the kind of speedup one might expect from running the algorithm on a supercomputer.

While we compare these packages separately, we recognize that there has been a considerable amount of ``borrowing back and forth'' between these and other projects. The notion of inferring causal graphs from conditional independence constraints goes back some time now and is used in all of these projects. The PC algorithm, as noted, originally appearing in an early version of Tetrad, is used in three of the projects, and algorithms derived from PC in two of them. The CPC algorithm, originally in Tetrad, was reimplemented in pcalg, along with some variants there. In bnlearn, the building blocks of other constraint-based algorithms are shuffled to generate CPDAGs from considerations of Markov blanket estimations and scoring algorithms. For GES, Chickering's (2002) algorithm was adjusted in each of pcalg and Tetrad; the specific formulation of the BIC score difference was adapted from ARGES \cite{nandy2015high} and is responsible for some of the speedup there.

Overall we find that for the PC-style algorithms, the algorithms of choice for accuracy are versions the ones based on CPC. However, these are dominated on recall by the GES-style algorithms, which among these options also include the fastest. This is perhaps surprising, since GES has had an  reputation among a number of researchers as being a slow algorithm, but both FGES and the pcalg implementation of GES prove this reputation to be undeserved; the algorithm offers many opportunities for optimization.

For future work, if algorithms are identified that should have been included in this comparison, we will include them and update the comparison. Beyond this condition, there are several more conditions to explore, as suggested above. We will take up as time permits.

\appendix
\section{Results Tables}

Tables of results follow. Here, 'Alg' is the algorithm as indexed in the text, 'vars' is the number of variables in each simulation, 'Deg' is the average degree of the graph, 'N' is the sample size, 'AP' is adjacency precision (as defined in the text), 'AR' adjacency recall, 'AHP' arrowhead precision, 'AHR' arrowhead recall, 'McAdj' Mathew's correlation for adjacencies, 'McArrow' Matthews correlation for orientations, and 'E' elapsed time in seconds. For hardware and software specifications, please see text. There is one table for each simulation; the conditions for the simulation are in the table and are common for all algorithms in each table. '*' in table indicates the value is undefined due to division by zero  and '-' means that the value is zero or that the simulations in that position in the table took longer than 10 minutes to complete. Each statistic is an average over 10 runs.

\begin{center}
\begin{longtable}{l l l l l l l l l l l l}
\hline
Alg & Vars & Deg & N & AP & AR & AHP & AHR & McAdj & McArrow & E \\ 
\hline
1 & 50 & 2 & 100 & 0.95 & 0.67 & 0.59 & 0.28 & 0.79 & 0.11 & 0.03 \\ 
2 & 50 & 2 & 100 & 0.99 & 0.56 & 0.55 & 0.2 & 0.73 & 0.06 & 0.04 \\ 
3 & 50 & 2 & 100 & 0.95 & 0.66 & 0.56 & 0.28 & 0.79 & 0.08 & 0.01 \\ 
4 & 50 & 2 & 100 & 0.99 & 0.55 & 0.54 & 0.19 & 0.73 & 0.04 & 0.01 \\ 
5 & 50 & 2 & 100 & 0.95 & 0.66 & 0.86 & 0.31 & 0.79 & 0.35 & 0.04 \\ 
6 & 50 & 2 & 100 & 0.99 & 0.55 & 0.86 & 0.21 & 0.73 & 0.27 & 0.01 \\ 
7 & 50 & 2 & 100 & 0.95 & 0.67 & 0.99 & 0.15 & 0.79 & 0.27 & 0.01 \\
\hline
\end{longtable}

\begin{longtable}{l l l l l l l l l l l l}
\hline
Alg & Vars & Deg & N & AP & AR & AHP & AHR & McAdj & McArrow & E \\ 
\hline
1 & 50 & 2 & 500 & 0.98 & 0.94 & 0.75 & 0.56 & 0.96 & 0.39 & 0.06 \\ 
2 & 50 & 2 & 500 & 1 & 0.91 & 0.73 & 0.5 & 0.95 & 0.34 & 0.02 \\ 
3 & 50 & 2 & 500 & 0.98 & 0.94 & 0.68 & 0.53 & 0.96 & 0.3 & 0.04 \\ 
4 & 50 & 2 & 500 & 1 & 0.9 & 0.67 & 0.48 & 0.95 & 0.26 & 0.04 \\ 
5 & 50 & 2 & 500 & 0.98 & 0.94 & 0.88 & 0.63 & 0.96 & 0.57 & 0.03 \\ 
6 & 50 & 2 & 500 & 1 & 0.9 & 0.91 & 0.59 & 0.95 & 0.57 & 0.02 \\ 
7 & 50 & 2 & 500 & 0.98 & 0.94 & 0.98 & 0.54 & 0.96 & 0.59 & 0.04 \\
\hline
\end{longtable}

\begin{longtable}{l l l l l l l l l l l l}
\hline
Alg & Vars & Deg & N & AP & AR & AHP & AHR & McAdj & McArrow & E \\ 
\hline1 & 50 & 2 & 1000 & 0.96 & 0.97 & 0.79 & 0.61 & 0.96 & 0.47 & 0.04 \\ 
2 & 50 & 2 & 1000 & 0.99 & 0.96 & 0.8 & 0.57 & 0.97 & 0.45 & 0.02 \\ 
3 & 50 & 2 & 1000 & 0.96 & 0.97 & 0.78 & 0.62 & 0.96 & 0.46 & 0.05 \\ 
4 & 50 & 2 & 1000 & 1 & 0.95 & 0.76 & 0.58 & 0.97 & 0.41 & 0.02 \\ 
5 & 50 & 2 & 1000 & 0.96 & 0.97 & 0.87 & 0.65 & 0.96 & 0.58 & 0.05 \\ 
6 & 50 & 2 & 1000 & 1 & 0.95 & 0.91 & 0.62 & 0.97 & 0.59 & 0.03 \\ 
7 & 50 & 2 & 1000 & 0.96 & 0.97 & 0.99 & 0.61 & 0.96 & 0.65 & 0.08 \\
\hline
\end{longtable}

\begin{longtable}{l l l l l l l l l l l l}
\hline
Alg & Vars & Deg & N & AP & AR & AHP & AHR & McAdj & McArrow & E \\ 
\hline1 & 50 & 4 & 100 & 0.98 & 0.48 & 0.57 & 0.21 & 0.67 & 0.06 & 9.70E-03 \\ 
2 & 50 & 4 & 100 & 0.99 & 0.39 & 0.52 & 0.15 & 0.6 & 0.01 & 4.70E-03 \\ 
3 & 50 & 4 & 100 & 0.99 & 0.46 & 0.51 & 0.19 & 0.66 & 0.01 & 9.30E-03 \\ 
4 & 50 & 4 & 100 & 0.99 & 0.37 & 0.48 & 0.13 & 0.59 & -0.02 & 4.20E-03 \\ 
5 & 50 & 4 & 100 & 0.99 & 0.46 & 0.78 & 0.24 & 0.66 & 0.24 & 0.01 \\ 
6 & 50 & 4 & 100 & 0.99 & 0.37 & 0.8 & 0.14 & 0.59 & 0.18 & 4.50E-03 \\ 
7 & 50 & 4 & 100 & 0.98 & 0.48 & 0.94 & 0.09 & 0.67 & 0.19 & 0.01 \\
\hline
\end{longtable}

\begin{longtable}{l l l l l l l l l l l l}
\hline
Alg & Vars & Deg & N & AP & AR & AHP & AHR & McAdj & McArrow & E \\ 
\hline1 & 50 & 4 & 500 & 0.98 & 0.79 & 0.68 & 0.49 & 0.87 & 0.27 & 0.07 \\ 
2 & 50 & 4 & 500 & 0.99 & 0.72 & 0.66 & 0.42 & 0.83 & 0.22 & 0.04 \\ 
3 & 50 & 4 & 500 & 0.98 & 0.78 & 0.65 & 0.46 & 0.87 & 0.22 & 0.08 \\ 
4 & 50 & 4 & 500 & 0.99 & 0.72 & 0.64 & 0.41 & 0.83 & 0.19 & 0.05 \\ 
5 & 50 & 4 & 500 & 0.98 & 0.78 & 0.82 & 0.58 & 0.87 & 0.48 & 0.11 \\ 
6 & 50 & 4 & 500 & 0.99 & 0.72 & 0.84 & 0.53 & 0.83 & 0.46 & 0.07 \\ 
7 & 50 & 4 & 500 & 0.98 & 0.79 & 0.93 & 0.47 & 0.87 & 0.5 & 0.21 \\
\hline
\end{longtable}

\begin{longtable}{l l l l l l l l l l l l}
\hline
Alg & Vars & Deg & N & AP & AR & AHP & AHR & McAdj & McArrow & E \\ 
\hline1 & 50 & 4 & 1000 & 0.98 & 0.86 & 0.7 & 0.57 & 0.91 & 0.34 & 0.12 \\ 
2 & 50 & 4 & 1000 & 0.99 & 0.83 & 0.68 & 0.52 & 0.9 & 0.28 & 0.06 \\ 
3 & 50 & 4 & 1000 & 0.99 & 0.85 & 0.66 & 0.53 & 0.91 & 0.26 & 0.15 \\ 
4 & 50 & 4 & 1000 & 0.99 & 0.82 & 0.64 & 0.48 & 0.9 & 0.22 & 0.14 \\ 
5 & 50 & 4 & 1000 & 0.99 & 0.85 & 0.82 & 0.67 & 0.91 & 0.53 & 0.23 \\ 
6 & 50 & 4 & 1000 & 0.99 & 0.82 & 0.83 & 0.64 & 0.9 & 0.52 & 0.18 \\ 
7 & 50 & 4 & 1000 & 0.98 & 0.86 & 0.95 & 0.63 & 0.91 & 0.64 & 0.49 \\
\hline
\end{longtable}

\begin{longtable}{l l l l l l l l l l l l}
\hline
Alg & Vars & Deg & N & AP & AR & AHP & AHR & McAdj & McArrow & E \\ 
\hline1 & 50 & 6 & 100 & 0.94 & 0.35 & 0.53 & 0.16 & 0.55 & 0.03 & 6.70E-03 \\ 
2 & 50 & 6 & 100 & 0.97 & 0.27 & 0.55 & 0.11 & 0.49 & 0.03 & 3.70E-03 \\ 
3 & 50 & 6 & 100 & 0.96 & 0.33 & 0.52 & 0.15 & 0.54 & 0.02 & 9.50E-03 \\ 
4 & 50 & 6 & 100 & 0.98 & 0.26 & 0.51 & 0.1 & 0.48 & 7.93E-03 & 6.30E-03 \\ 
5 & 50 & 6 & 100 & 0.96 & 0.33 & 0.66 & 0.18 & 0.54 & 0.13 & 0.01 \\ 
6 & 50 & 6 & 100 & 0.98 & 0.26 & 0.74 & 0.1 & 0.48 & 0.12 & 5.90E-03 \\ 
7 & 50 & 6 & 100 & 0.94 & 0.35 & 0.83 & 0.09 & 0.55 & 0.15 & 0.01 \\
\hline
\end{longtable}

\begin{longtable}{l l l l l l l l l l l l}
\hline
Alg & Vars & Deg & N & AP & AR & AHP & AHR & McAdj & McArrow & E \\ 
\hline1 & 50 & 6 & 500 & 0.94 & 0.59 & 0.61 & 0.35 & 0.72 & 0.15 & 0.08 \\ 
2 & 50 & 6 & 500 & 0.95 & 0.54 & 0.61 & 0.32 & 0.69 & 0.13 & 0.05 \\ 
3 & 50 & 6 & 500 & 0.96 & 0.57 & 0.59 & 0.32 & 0.72 & 0.11 & 0.17 \\ 
4 & 50 & 6 & 500 & 0.97 & 0.53 & 0.57 & 0.28 & 0.69 & 0.09 & 0.07 \\ 
5 & 50 & 6 & 500 & 0.96 & 0.57 & 0.74 & 0.42 & 0.72 & 0.31 & 0.16 \\ 
6 & 50 & 6 & 500 & 0.97 & 0.53 & 0.72 & 0.37 & 0.69 & 0.27 & 0.14 \\ 
7 & 50 & 6 & 500 & 0.94 & 0.59 & 0.87 & 0.35 & 0.72 & 0.37 & 0.28 \\
\hline
\end{longtable}

\begin{longtable}{l l l l l l l l l l l l}
\hline
Alg & Vars & Deg & N & AP & AR & AHP & AHR & McAdj & McArrow & E \\ 
\hline1 & 50 & 6 & 1000 & 0.93 & 0.63 & 0.6 & 0.39 & 0.74 & 0.15 & 0.22 \\ 
2 & 50 & 6 & 1000 & 0.93 & 0.59 & 0.57 & 0.34 & 0.72 & 0.1 & 0.12 \\ 
3 & 50 & 6 & 1000 & 0.95 & 0.61 & 0.57 & 0.35 & 0.74 & 0.1 & 0.37 \\ 
4 & 50 & 6 & 1000 & 0.95 & 0.57 & 0.54 & 0.31 & 0.71 & 0.06 & 0.16 \\ 
5 & 50 & 6 & 1000 & 0.95 & 0.61 & 0.72 & 0.45 & 0.74 & 0.3 & 0.36 \\ 
6 & 50 & 6 & 1000 & 0.95 & 0.57 & 0.72 & 0.41 & 0.71 & 0.28 & 0.26 \\ 
7 & 50 & 6 & 1000 & 0.93 & 0.63 & 0.85 & 0.41 & 0.74 & 0.4 & 0.77 \\
\hline
\end{longtable}

\begin{longtable}{l l l l l l l l l l l l}
\hline
Alg & Vars & Deg & N & AP & AR & AHP & AHR & McAdj & McArrow & E \\ 
\hline1 & 100 & 2 & 100 & 0.89 & 0.64 & 0.52 & 0.27 & 0.75 & 0.03 & 0.01 \\ 
2 & 100 & 2 & 100 & 0.97 & 0.54 & 0.56 & 0.2 & 0.72 & 0.06 & 0.02 \\ 
3 & 100 & 2 & 100 & 0.91 & 0.64 & 0.48 & 0.24 & 0.76 & -8.49E-03 & 5.30E-03 \\ 
4 & 100 & 2 & 100 & 0.98 & 0.53 & 0.56 & 0.2 & 0.72 & 0.05 & 4.40E-03 \\ 
5 & 100 & 2 & 100 & 0.91 & 0.64 & 0.76 & 0.3 & 0.76 & 0.25 & 7.60E-03 \\ 
6 & 100 & 2 & 100 & 0.98 & 0.53 & 0.84 & 0.21 & 0.72 & 0.26 & 5.70E-03 \\ 
7 & 100 & 2 & 100 & 0.89 & 0.64 & 0.97 & 0.13 & 0.75 & 0.25 & 0.02 \\
\hline
\end{longtable}

\begin{longtable}{l l l l l l l l l l l l}
\hline
Alg & Vars & Deg & N & AP & AR & AHP & AHR & McAdj & McArrow & E \\ 
\hline1 & 100 & 2 & 500 & 0.92 & 0.95 & 0.72 & 0.58 & 0.93 & 0.38 & 0.05 \\ 
2 & 100 & 2 & 500 & 1 & 0.91 & 0.77 & 0.53 & 0.95 & 0.39 & 0.03 \\ 
3 & 100 & 2 & 500 & 0.93 & 0.94 & 0.66 & 0.54 & 0.93 & 0.29 & 0.04 \\ 
4 & 100 & 2 & 500 & 1 & 0.91 & 0.7 & 0.5 & 0.95 & 0.3 & 0.02 \\ 
5 & 100 & 2 & 500 & 0.93 & 0.94 & 0.83 & 0.62 & 0.93 & 0.52 & 0.04 \\ 
6 & 100 & 2 & 500 & 1 & 0.91 & 0.91 & 0.58 & 0.95 & 0.57 & 0.05 \\ 
7 & 100 & 2 & 500 & 0.92 & 0.95 & 0.99 & 0.53 & 0.93 & 0.59 & 0.07 \\
\hline
\end{longtable}

\begin{longtable}{l l l l l l l l l l l l}
\hline
Alg & Vars & Deg & N & AP & AR & AHP & AHR & McAdj & McArrow & E \\ 
\hline1 & 100 & 2 & 1000 & 0.94 & 0.99 & 0.77 & 0.64 & 0.96 & 0.46 & 0.06 \\ 
2 & 100 & 2 & 1000 & 0.99 & 0.97 & 0.81 & 0.63 & 0.98 & 0.5 & 0.04 \\ 
3 & 100 & 2 & 1000 & 0.94 & 0.99 & 0.73 & 0.62 & 0.96 & 0.41 & 0.1 \\ 
4 & 100 & 2 & 1000 & 0.99 & 0.97 & 0.77 & 0.6 & 0.98 & 0.43 & 0.04 \\ 
5 & 100 & 2 & 1000 & 0.94 & 0.99 & 0.89 & 0.7 & 0.96 & 0.63 & 0.09 \\ 
6 & 100 & 2 & 1000 & 0.99 & 0.97 & 0.95 & 0.68 & 0.98 & 0.67 & 0.05 \\ 
7 & 100 & 2 & 1000 & 0.94 & 0.99 & 1 & 0.63 & 0.96 & 0.67 & 0.15 \\
\hline
\end{longtable}

\begin{longtable}{l l l l l l l l l l l l}
\hline
Alg & Vars & Deg & N & AP & AR & AHP & AHR & McAdj & McArrow & E \\ 
\hline1 & 100 & 4 & 100 & 0.96 & 0.5 & 0.59 & 0.25 & 0.69 & 0.09 & 0.03 \\ 
2 & 100 & 4 & 100 & 0.99 & 0.38 & 0.56 & 0.16 & 0.61 & 0.05 & 0.02 \\ 
3 & 100 & 4 & 100 & 0.97 & 0.49 & 0.52 & 0.22 & 0.68 & 0.02 & 0.02 \\ 
4 & 100 & 4 & 100 & 0.99 & 0.37 & 0.49 & 0.14 & 0.6 & -0.01 & 5.90E-03 \\ 
5 & 100 & 4 & 100 & 0.97 & 0.49 & 0.78 & 0.26 & 0.68 & 0.25 & 0.02 \\ 
6 & 100 & 4 & 100 & 0.99 & 0.37 & 0.75 & 0.15 & 0.6 & 0.16 & 8.70E-03 \\ 
7 & 100 & 4 & 100 & 0.96 & 0.5 & 0.97 & 0.11 & 0.69 & 0.22 & 0.03 \\
\hline
\end{longtable}

\begin{longtable}{l l l l l l l l l l l l}
\hline
Alg & Vars & Deg & N & AP & AR & AHP & AHR & McAdj & McArrow & E \\ 
\hline1 & 100 & 4 & 500 & 0.98 & 0.83 & 0.7 & 0.55 & 0.9 & 0.33 & 0.12 \\ 
2 & 100 & 4 & 500 & 0.99 & 0.78 & 0.68 & 0.48 & 0.87 & 0.27 & 0.06 \\ 
3 & 100 & 4 & 500 & 0.98 & 0.82 & 0.63 & 0.49 & 0.9 & 0.22 & 0.17 \\ 
4 & 100 & 4 & 500 & 0.99 & 0.77 & 0.62 & 0.44 & 0.87 & 0.18 & 0.08 \\ 
5 & 100 & 4 & 500 & 0.98 & 0.82 & 0.85 & 0.66 & 0.9 & 0.56 & 0.28 \\ 
6 & 100 & 4 & 500 & 0.99 & 0.77 & 0.86 & 0.61 & 0.87 & 0.53 & 0.22 \\ 
7 & 100 & 4 & 500 & 0.98 & 0.83 & 0.96 & 0.55 & 0.9 & 0.58 & 0.49 \\
\hline
\end{longtable}

\begin{longtable}{l l l l l l l l l l l l}
\hline
Alg & Vars & Deg & N & AP & AR & AHP & AHR & McAdj & McArrow & E \\ 
\hline1 & 100 & 4 & 1000 & 0.99 & 0.88 & 0.73 & 0.6 & 0.93 & 0.39 & 0.25 \\ 
2 & 100 & 4 & 1000 & 1 & 0.85 & 0.74 & 0.58 & 0.92 & 0.39 & 0.18 \\ 
3 & 100 & 4 & 1000 & 0.99 & 0.87 & 0.68 & 0.55 & 0.93 & 0.3 & 0.47 \\ 
4 & 100 & 4 & 1000 & 1 & 0.84 & 0.66 & 0.52 & 0.91 & 0.26 & 0.21 \\ 
5 & 100 & 4 & 1000 & 0.99 & 0.87 & 0.9 & 0.72 & 0.93 & 0.65 & 0.6 \\ 
6 & 100 & 4 & 1000 & 1 & 0.84 & 0.9 & 0.68 & 0.91 & 0.62 & 0.49 \\ 
7 & 100 & 4 & 1000 & 0.99 & 0.88 & 0.97 & 0.68 & 0.93 & 0.69 & 1.28 \\
\hline
\end{longtable}

\begin{longtable}{l l l l l l l l l l l l}
\hline
Alg & Vars & Deg & N & AP & AR & AHP & AHR & McAdj & McArrow & E \\ 
\hline1 & 100 & 6 & 100 & 0.95 & 0.37 & 0.55 & 0.18 & 0.58 & 0.04 & 0.02 \\ 
2 & 100 & 6 & 100 & 0.98 & 0.3 & 0.52 & 0.12 & 0.52 & 0.02 & 0.03 \\ 
3 & 100 & 6 & 100 & 0.96 & 0.36 & 0.5 & 0.16 & 0.58 & -1.16E-05 & 0.03 \\ 
4 & 100 & 6 & 100 & 0.98 & 0.28 & 0.46 & 0.11 & 0.51 & -0.03 & 0.02 \\ 
5 & 100 & 6 & 100 & 0.96 & 0.36 & 0.71 & 0.21 & 0.58 & 0.17 & 0.03 \\ 
6 & 100 & 6 & 100 & 0.98 & 0.28 & 0.72 & 0.12 & 0.51 & 0.14 & 0.02 \\ 
7 & 100 & 6 & 100 & 0.95 & 0.37 & 0.89 & 0.07 & 0.58 & 0.16 & 0.03 \\
\hline
\end{longtable}

\begin{longtable}{l l l l l l l l l l l l}
\hline
Alg & Vars & Deg & N & AP & AR & AHP & AHR & McAdj & McArrow & E \\ 
\hline1 & 100 & 6 & 500 & 0.96 & 0.65 & 0.64 & 0.4 & 0.78 & 0.2 & 0.24 \\ 
2 & 100 & 6 & 500 & 0.97 & 0.59 & 0.65 & 0.37 & 0.75 & 0.2 & 0.16 \\ 
3 & 100 & 6 & 500 & 0.97 & 0.64 & 0.6 & 0.37 & 0.78 & 0.14 & 0.41 \\ 
4 & 100 & 6 & 500 & 0.98 & 0.58 & 0.59 & 0.33 & 0.74 & 0.12 & 0.15 \\ 
5 & 100 & 6 & 500 & 0.97 & 0.64 & 0.77 & 0.48 & 0.78 & 0.37 & 0.54 \\ 
6 & 100 & 6 & 500 & 0.98 & 0.58 & 0.8 & 0.44 & 0.74 & 0.38 & 0.3 \\ 
7 & 100 & 6 & 500 & 0.96 & 0.65 & 0.92 & 0.41 & 0.78 & 0.46 & 0.98 \\
\hline
\end{longtable}

\begin{longtable}{l l l l l l l l l l l l}
\hline
Alg & Vars & Deg & N & AP & AR & AHP & AHR & McAdj & McArrow & E \\ 
\hline1 & 100 & 6 & 1000 & 0.97 & 0.72 & 0.64 & 0.46 & 0.83 & 0.22 & 0.66 \\ 
2 & 100 & 6 & 1000 & 0.98 & 0.68 & 0.65 & 0.43 & 0.81 & 0.21 & 0.33 \\ 
3 & 100 & 6 & 1000 & 0.97 & 0.71 & 0.6 & 0.42 & 0.82 & 0.15 & 1.24 \\ 
4 & 100 & 6 & 1000 & 0.98 & 0.67 & 0.59 & 0.38 & 0.8 & 0.13 & 0.69 \\ 
5 & 100 & 6 & 1000 & 0.97 & 0.71 & 0.81 & 0.56 & 0.82 & 0.45 & 1.51 \\ 
6 & 100 & 6 & 1000 & 0.98 & 0.67 & 0.82 & 0.53 & 0.8 & 0.45 & 1.06 \\ 
7 & 100 & 6 & 1000 & 0.97 & 0.72 & 0.93 & 0.53 & 0.83 & 0.55 & 3.23 \\
\hline
\end{longtable}

\begin{longtable}{l l l l l l l l l l l l}
\hline
Alg & Vars & Deg & N & AP & AR & AHP & AHR & McAdj & McArrow & E \\ 
\hline1 & 500 & 2 & 100 & 0.74 & 0.66 & 0.42 & 0.29 & 0.69 & -0.05 & 1.22 \\ 
2 & 500 & 2 & 100 & 0.93 & 0.55 & 0.52 & 0.18 & 0.71 & 0.03 & 1.22 \\ 
3 & 500 & 2 & 100 & 0.79 & 0.65 & 0.43 & 0.27 & 0.71 & -0.04 & 0.18 \\ 
4 & 500 & 2 & 100 & 0.95 & 0.54 & 0.5 & 0.17 & 0.71 & 7.10E-03 & 0.06 \\ 
5 & 500 & 2 & 100 & 0.79 & 0.65 & 0.63 & 0.3 & 0.71 & 0.17 & 0.12 \\ 
6 & 500 & 2 & 100 & 0.95 & 0.54 & 0.84 & 0.19 & 0.71 & 0.24 & 0.05 \\ 
7 & 500 & 2 & 100 & 0.74 & 0.66 & 0.96 & 0.12 & 0.69 & 0.24 & 1.6 \\
\hline
\end{longtable}

\begin{longtable}{l l l l l l l l l l l l}
\hline
Alg & Vars & Deg & N & AP & AR & AHP & AHR & McAdj & McArrow & E \\ 
\hline1 & 500 & 2 & 500 & 0.78 & 0.95 & 0.53 & 0.55 & 0.86 & 0.16 & 1.56 \\ 
2 & 500 & 2 & 500 & 0.97 & 0.91 & 0.71 & 0.52 & 0.94 & 0.33 & 1.23 \\ 
3 & 500 & 2 & 500 & 0.8 & 0.95 & 0.53 & 0.54 & 0.87 & 0.15 & 0.39 \\ 
4 & 500 & 2 & 500 & 0.98 & 0.91 & 0.68 & 0.5 & 0.94 & 0.28 & 0.18 \\ 
5 & 500 & 2 & 500 & 0.8 & 0.95 & 0.7 & 0.68 & 0.87 & 0.44 & 0.46 \\ 
6 & 500 & 2 & 500 & 0.98 & 0.91 & 0.92 & 0.63 & 0.94 & 0.61 & 0.25 \\ 
7 & 500 & 2 & 500 & 0.78 & 0.95 & 0.99 & 0.55 & 0.86 & 0.61 & 1.66 \\
\hline
\end{longtable}

\begin{longtable}{l l l l l l l l l l l l}
\hline
Alg & Vars & Deg & N & AP & AR & AHP & AHR & McAdj & McArrow & E \\ 
\hline1 & 500 & 2 & 1000 & 0.77 & 0.99 & 0.57 & 0.62 & 0.87 & 0.25 & 2.02 \\ 
2 & 500 & 2 & 1000 & 0.98 & 0.98 & 0.8 & 0.62 & 0.98 & 0.48 & 1.41 \\ 
3 & 500 & 2 & 1000 & 0.78 & 0.99 & 0.56 & 0.6 & 0.88 & 0.22 & 0.78 \\ 
4 & 500 & 2 & 1000 & 0.98 & 0.98 & 0.76 & 0.6 & 0.98 & 0.43 & 0.37 \\ 
5 & 500 & 2 & 1000 & 0.78 & 0.99 & 0.71 & 0.73 & 0.88 & 0.48 & 0.89 \\ 
6 & 500 & 2 & 1000 & 0.98 & 0.98 & 0.95 & 0.69 & 0.98 & 0.68 & 0.53 \\ 
7 & 500 & 2 & 1000 & 0.77 & 0.99 & 0.99 & 0.66 & 0.87 & 0.69 & 2.36 \\
\hline
\end{longtable}

\begin{longtable}{l l l l l l l l l l l l}
\hline
Alg & Vars & Deg & N & AP & AR & AHP & AHR & McAdj & McArrow & E \\ 
\hline1 & 500 & 4 & 100 & 0.91 & 0.5 & 0.54 & 0.24 & 0.67 & 0.05 & 1.17 \\ 
2 & 500 & 4 & 100 & 0.98 & 0.4 & 0.56 & 0.16 & 0.62 & 0.05 & 1.36 \\ 
3 & 500 & 4 & 100 & 0.93 & 0.49 & 0.5 & 0.22 & 0.67 & 7.30E-03 & 0.15 \\ 
4 & 500 & 4 & 100 & 0.99 & 0.38 & 0.52 & 0.15 & 0.62 & 0.02 & 0.1 \\ 
5 & 500 & 4 & 100 & 0.93 & 0.49 & 0.76 & 0.3 & 0.67 & 0.26 & 0.18 \\ 
6 & 500 & 4 & 100 & 0.99 & 0.38 & 0.85 & 0.17 & 0.62 & 0.23 & 0.11 \\ 
7 & 500 & 4 & 100 & 0.91 & 0.5 & 0.97 & 0.13 & 0.67 & 0.25 & 1.14 \\
\hline
\end{longtable}

\begin{longtable}{l l l l l l l l l l l l}
\hline
Alg & Vars & Deg & N & AP & AR & AHP & AHR & McAdj & McArrow & E \\ 
\hline1 & 500 & 4 & 500 & 0.96 & 0.87 & 0.68 & 0.58 & 0.91 & 0.32 & 2.16 \\ 
2 & 500 & 4 & 500 & 1 & 0.81 & 0.69 & 0.51 & 0.89 & 0.29 & 1.76 \\ 
3 & 500 & 4 & 500 & 0.96 & 0.87 & 0.63 & 0.53 & 0.91 & 0.23 & 1.46 \\ 
4 & 500 & 4 & 500 & 1 & 0.8 & 0.63 & 0.47 & 0.89 & 0.2 & 0.63 \\ 
5 & 500 & 4 & 500 & 0.96 & 0.87 & 0.88 & 0.72 & 0.91 & 0.64 & 2.33 \\ 
6 & 500 & 4 & 500 & 1 & 0.8 & 0.93 & 0.66 & 0.89 & 0.64 & 1.41 \\ 
7 & 500 & 4 & 500 & 0.96 & 0.87 & 0.99 & 0.6 & 0.91 & 0.65 & 5.91 \\
\hline
\end{longtable}

\begin{longtable}{l l l l l l l l l l l l}
\hline
Alg & Vars & Deg & N & AP & AR & AHP & AHR & McAdj & McArrow & E \\ 
\hline1 & 500 & 4 & 1000 & 0.97 & 0.94 & 0.74 & 0.68 & 0.96 & 0.44 & 3.82 \\ 
2 & 500 & 4 & 1000 & 1 & 0.91 & 0.75 & 0.64 & 0.95 & 0.42 & 2.62 \\ 
3 & 500 & 4 & 1000 & 0.97 & 0.94 & 0.69 & 0.63 & 0.96 & 0.36 & 4.33 \\ 
4 & 500 & 4 & 1000 & 1 & 0.91 & 0.69 & 0.59 & 0.95 & 0.33 & 2.32 \\ 
5 & 500 & 4 & 1000 & 0.97 & 0.94 & 0.91 & 0.82 & 0.96 & 0.75 & 7.54 \\ 
6 & 500 & 4 & 1000 & 1 & 0.91 & 0.95 & 0.79 & 0.95 & 0.76 & 5.45 \\ 
7 & 500 & 4 & 1000 & 0.97 & 0.94 & 1 & 0.75 & 0.96 & 0.77 & 19.59 \\
\hline
\end{longtable}

\begin{longtable}{l l l l l l l l l l l l}
\hline
Alg & Vars & Deg & N & AP & AR & AHP & AHR & McAdj & McArrow & E \\ 
\hline1 & 500 & 6 & 100 & 0.94 & 0.39 & 0.56 & 0.19 & 0.6 & 0.06 & 1.22 \\ 
2 & 500 & 6 & 100 & 0.99 & 0.3 & 0.55 & 0.13 & 0.54 & 0.03 & 1.24 \\ 
3 & 500 & 6 & 100 & 0.96 & 0.38 & 0.52 & 0.17 & 0.6 & 0.02 & 0.26 \\ 
4 & 500 & 6 & 100 & 0.99 & 0.28 & 0.5 & 0.11 & 0.53 & 8.32E-04 & 0.16 \\ 
5 & 500 & 6 & 100 & 0.96 & 0.38 & 0.76 & 0.24 & 0.6 & 0.22 & 0.28 \\ 
6 & 500 & 6 & 100 & 0.99 & 0.28 & 0.83 & 0.13 & 0.53 & 0.19 & 0.14 \\ 
7 & 500 & 6 & 100 & 0.94 & 0.39 & 0.98 & 0.09 & 0.6 & 0.21 & 1.33 \\
\hline
\end{longtable}

\begin{longtable}{l l l l l l l l l l l l}
\hline
Alg & Vars & Deg & N & AP & AR & AHP & AHR & McAdj & McArrow & E \\ 
\hline1 & 500 & 6 & 500 & 0.99 & 0.75 & 0.65 & 0.48 & 0.86 & 0.24 & 3.86 \\ 
2 & 500 & 6 & 500 & 1 & 0.68 & 0.65 & 0.42 & 0.82 & 0.21 & 2.4 \\ 
3 & 500 & 6 & 500 & 0.99 & 0.74 & 0.6 & 0.43 & 0.86 & 0.15 & 4.57 \\ 
4 & 500 & 6 & 500 & 1 & 0.67 & 0.59 & 0.38 & 0.82 & 0.13 & 1.87 \\ 
5 & 500 & 6 & 500 & 0.99 & 0.74 & 0.88 & 0.64 & 0.86 & 0.57 & 7.39 \\ 
6 & 500 & 6 & 500 & 1 & 0.67 & 0.9 & 0.58 & 0.82 & 0.55 & 3.92 \\ 
7 & 500 & 6 & 500 & 0.99 & 0.75 & 0.99 & 0.52 & 0.86 & 0.58 & 16.45 \\
\hline
\end{longtable}

\begin{longtable}{l l l l l l l l l l l l}
\hline
Alg & Vars & Deg & N & AP & AR & AHP & AHR & McAdj & McArrow & E \\ 
\hline1 & 500 & 6 & 1000 & 0.99 & 0.85 & 0.69 & 0.57 & 0.91 & 0.32 & 11.13 \\ 
2 & 500 & 6 & 1000 & 1 & 0.8 & 0.68 & 0.53 & 0.89 & 0.29 & 5.79 \\ 
3 & 500 & 6 & 1000 & 0.99 & 0.84 & 0.64 & 0.53 & 0.91 & 0.23 & 15.7 \\ 
4 & 500 & 6 & 1000 & 1 & 0.79 & 0.62 & 0.48 & 0.89 & 0.2 & 6.37 \\ 
5 & 500 & 6 & 1000 & 0.99 & 0.84 & 0.9 & 0.74 & 0.91 & 0.67 & 31.96 \\ 
6 & 500 & 6 & 1000 & 1 & 0.79 & 0.91 & 0.7 & 0.89 & 0.65 & 17.17 \\ 
7 & 500 & 6 & 1000 & 0.99 & 0.85 & 0.99 & 0.68 & 0.91 & 0.7 & 81.91 \\
\hline
\end{longtable}

\end{center}

\section*{References}
\bibliographystyle{plain}
\bibliography{sample}

\end{document}